\documentclass[10pt,twocolumn,letterpaper]{article}

\usepackage{cvpr}
\usepackage{times}
\usepackage{epsfig}
\usepackage{graphicx}
\usepackage{amsmath}
\usepackage{amssymb}

\usepackage{tabularx}
\usepackage{multirow}

\usepackage{subcaption}
\usepackage{adjustbox}

\newcommand*\rot{\rotatebox{90}}

\usepackage[pagebackref=true,breaklinks=true,letterpaper=true,colorlinks,bookmarks=false]{hyperref}

\cvprfinalcopy 

\def\cvprPaperID{1931} 

\ifcvprfinal\fi

\begin{document}

\title{Spatially-Adaptive Filter Units for Deep Neural Networks}


\author{Domen Tabernik$^1$, Matej kristan$^1$ and Ale\v{s} Leonardis$^{1,2}$\\
$^1$Faculty of Computer and Information Science, University of Ljubljana, Ljubljana, Slovenia\\
$^2$CN-CR Centre, School of Computer Science, University of Birmingham, Birmingham, UK\\
{\tt\small \{domen.tabernik,matej.kristan\}@fri.uni-lj.si}\\
{\tt\small a.leonardis@cs.bham.ac.uk}
}

\maketitle

\begin{abstract}
 
Classical deep convolutional networks increase receptive field size by either gradual resolution reduction or application of hand-crafted dilated convolutions to prevent increase in the number of parameters. In this paper we propose a novel displaced aggregation unit (DAU) that does not require hand-crafting. In contrast to classical filters with units (pixels) placed on a fixed regular grid, the displacement of the DAUs are learned, which enables filters to spatially-adapt their receptive field to a given problem. We extensively demonstrate the strength of DAUs on a classification and semantic segmentation tasks. Compared to ConvNets with regular filter, ConvNets with DAUs achieve comparable performance at faster convergence and up to 3-times reduction in parameters. Furthermore, DAUs allow us to study deep networks from novel perspectives. We study spatial distributions of DAU filters and analyze the number of parameters allocated for spatial coverage in a filter.

\end{abstract}

\section{Introduction}

Deep convolutional neural networks (ConvNet)~\cite{He2015a,Xie2016c,Redmon2016,He2017} have become prevalent in visual feature learning. The integral part of these approaches are convolutional filters. In combination with other layers, the definition of the filter directly influences the kind of features a network can capture. Current state-of-the-art ConvNets define filters as  rectangular windows of weights where each learnable unit is a single pixel value in the filter. 

\begin{figure}
\includegraphics[width=\linewidth]{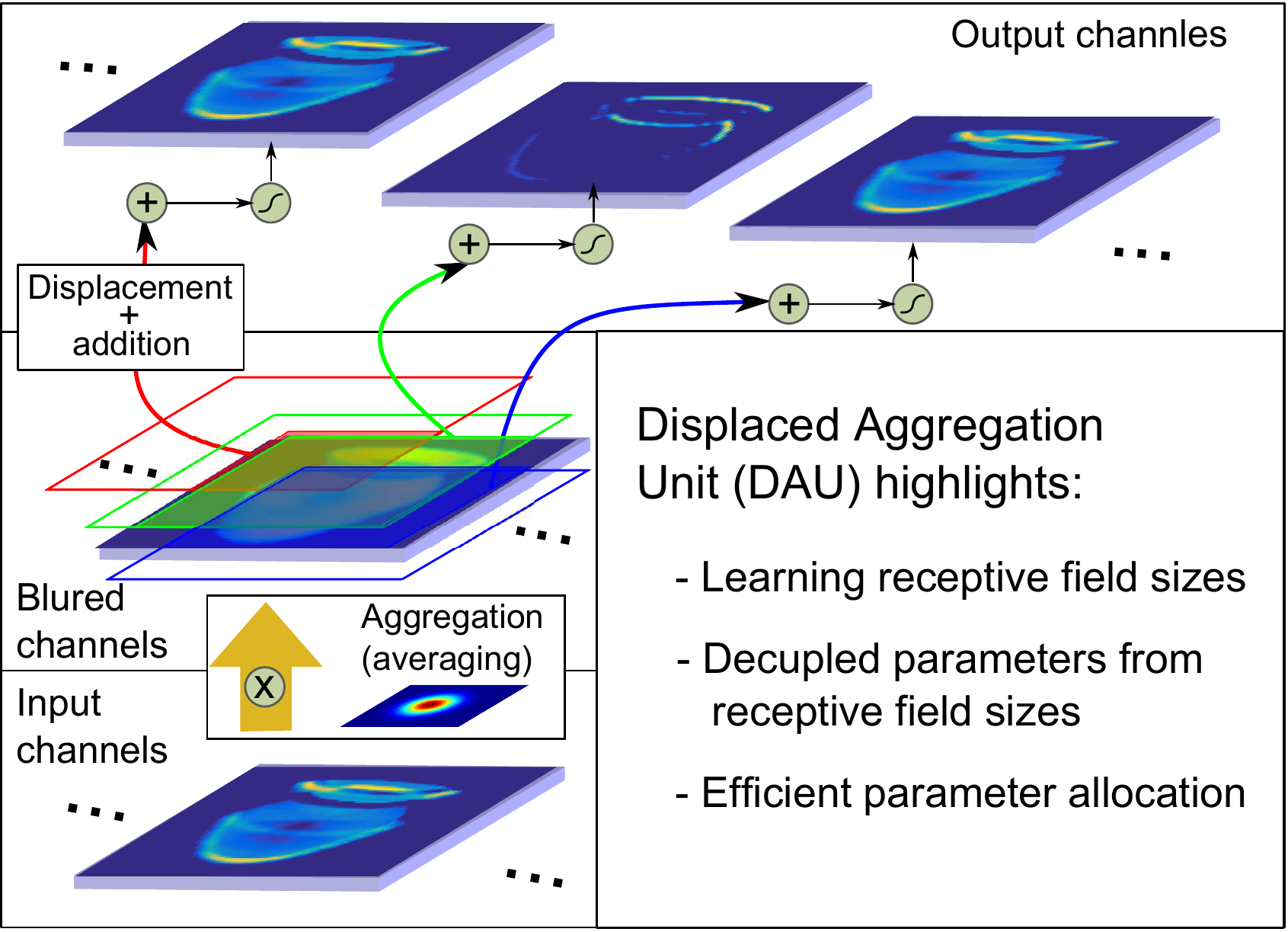}
\vspace*{-1.5em}
\caption{The displaced aggregation units (DAUs) afford efficient implementation. Convolution of a feature channel with a filter composed of several DAUs is implemented as  blurring by a single Gaussian and subsampling at learned displacements.\label{fig:intro}}
\vspace*{-0.5em}
\end{figure}

An important hyperparameter of the filters is their size, which is directly related to the number of free parameters in ConvNets. Large filters are avoided in the interest of keeping this number low and reducing overfitting. On the other hand, feature expressiveness improves with increased receptive fields~\cite{Chatfield2014}. Classification networks thus apply small filters and implicitly increase the receptive field size by gradually reducing resolution via pooling layers and increased depth~\cite{Simonyan2015}. But in dense prediction problems like segmentation~\cite{Long2015,Chen2014}, sufficient resolution is required for accurate localization of segmentation boundaries. Thus large receptive fields have to explicitly be accounted for without resolution loss~\cite{Chen2017,Yu2015,Yu2017}.

Increasing a receptive field without sacrificing resolution is addressed by dilated (atrous) convolution~\cite{Yu2015,Yu2017}. This approach increases the kernel receptive field by spreading out (dilating) the positions of the kernel sampling units (i.e., pixels). Large dilations significantly violate Nyquist theorem~\cite{Amidror2013}, resulting in griding artifacts~\cite{Yu2017}. Mitigation of these requires additional convolutional layers with progressively smaller dilations. The dilation factors are another hyperparameter that is manually tuned. To alleviate manual specification to some extent,~\cite{Chen2014,Chen2017} propose to use several pre-selected dilation factors and achieve excellent results. 
  
We define a convolutional filter as a mixture of several displaced aggregation units which is a generalization of the convolutional layers typically used for classification and those for segmentation. In contrast to a standard filter, the aggregation unit is not a single pixel, but a locally averaged response. In our implementation, a Gaussian with a fixed variance is used for averaging. In contrast to standard ConvNets, our units are not positioned on a regular grid. Their displacements are adapted during learning, thus the receptive field size of each filter is tuned separately. This allows for large or small receptive fields without changing the number of parameters, facilitating automatic and efficient allocation of parameters.
 
Our major contribution is an efficient formulation of the {\em displaced aggregation units} (DAU) filter with sub-pixel displacements, which allows practical use in deep architectures (see Fig.~\ref{fig:intro}). The DAUs remove the requirement for hand-crafted dilation without modification of other layers, decouple the parameter count from the receptive field size and do not suffer from gridding effects. Backpropagation is derived for all parameters and the new layers are implemented in standard ConvNet package with low-level CUDA procedures~\cite{Jia2014}. Our secondary contributions are analyses that have not been possible with the existing networks. We demonstrate that a {\em single} type of DAU-based filters achieve comparable performance to standard ConvNets on classification as well as dilated ConvNets on a segmentation task. We perform analysis of the dilation patterns required for accurate segmentation by recording the distributions of the \textit{learned} displacements in DAUs. Our parameter study demonstrates that using only few DAUs per filter already results in excellent performance. Our tests also show that DAUs allow comparable performance to classical ConvNets at almost 3-fold reduction of the learned parameters in convolutional layers. 
 
The remainder of this paper is structured as follows: in Sec.~\ref{sec:related-work} we review most closely related works, we describe DAU in Sec.~\ref{sec:method} and evaluate our model in Sec.~\ref{sec:eval}. In Sec.~\ref{sec:dau-analysis} we present a comprehensive study of DAU filter displacements and conclude with a discussion in Sec.~\ref{sec:conclusion}.

\section{Related work\label{sec:related-work}}

Receptive field has been considered as an important factor for deep networks in several related works~\cite{Luo2016,Chen2017}. Luo et al.~\cite{Luo2016} measured an effective receptive field in convolutional neural networks and observed it increases as the network learns. They suggest an architectural change that foregos a rectangular windows of weights for a sparsely connected units. However, they do not show how this can be implemented. Our proposed approach is in direct alignment with their suggested changes as our displaced aggregation units are a direct realization of their suggested sparsely connected units.

The importance of deforming filter units has also been indicated by recent work of Dai et al.~\cite{Dai2017} and Jeon et al.~\cite{Jeon2017}. Dai et al.~\cite{Dai2017} implemented spatial deformation of features with deformable convolutional networks. They explicitly learn feature displacement but learn them on a per-pixel location basis for input activation map and share them between all channels and features. Our model instead learns different displacements for different channels and features, and shares them over all pixel locations in the input activation map. This makes our model complementary to deformable convolutions.  Jeon et al.~\cite{Jeon2017}, on the other hand, apply deformation on filter units similarly as we do. They use bilinear interpolation similar to ours to get displacements at a sub-pixel accuracy but they apply them to $3\times3$ filters. However, they do not learn different offsets for each channel but apply the same offset across all channels and features. This prevents them from decreeing their parameter count as they still use 9 units per filter. We show that significantly less units are needed.

Works by Luan et al.~\cite{Luan2017} and Jacobsen et al.~\cite{Jacobsen2016} changed filter definition using different parametrization techniques.  Both decompose filter units into a linear combination of edge filters. They show a reduction in parameters per filter but their models do not provide displacements of filter units to arbitrary values. Their models have a fixed receptive fields defined as a hyperparameter and cannot be learned as ours. This also prevents any further analysis on distribution of displacements and receptive field sizes which is possible with our model.

Our model also uses concepts for filter parametrization similar to Tabernik et al.~\cite{Tabernik2016a} but differs significantly in their design. The model by Tabernik et al.~\cite{Tabernik2016a} is limited to only small scale networks and implements only a shallow network with two convolutional layers due to  inefficient parametrization design. Our proposed model enjoys an efficient parametrization and we apply it to larger problems using deeper networks.


\section{Displaced aggregation units (DAU)~\label{sec:method}}


The activation map of the $i$-th feature (input into the current layer of neurons), is defined in the standard ConvNets as
 \begin{equation}
 	Y_{i} = f(\sum\nolimits_s W_{s}\ast X_{s}+b_{s}),
 \end{equation}
where $b_s$ is a bias, $\ast$ is a convolution operation between the input  map $X_s$ and the filter $W_{s}$, and $f(\cdot)$ is a non-linear function, such as ReLU or sigmoid~\cite{LeCun1998}. 

We define the filters $W_{s}$ as mixtures of localized aggregated feature responses from the input feature map. We choose Gaussians as an analytic form of aggregation units and compactly write filter as $W_{s}=\sum\nolimits_k w_{k}G(\boldsymbol{\mu}_k; \sigma)$, where the unit displacement and aggregation range are specified by the mean $\boldsymbol{\mu}_k$ and variance $\sigma^2$, respectively, and $w_{k}$ is the input amplification factor. With the exception of variance $\sigma^2$, the parameters $\boldsymbol{\mu}_k$ and $w_{k}$ are unique for each output feature $i$ and channel $s$, however, we omit this in notation for clarity. Note that mixtures of Gaussians have recently been explored as potential filters in~\cite{Tabernik2016a}. But due to the computational complexity of adapting all parameters, the approach was not feasible beyond a two-layer architecture.

In our preliminary study we noticed that while the unit locations play a crucial role in the shallow network performance, the variances do not. We thus make all variances in the Gaussians equal and fixed to a selected value, making the unit aggregation perimeter a single hyperparameter.

\subsection{Inference with DAU~\label{sec:inference}}

The DAUs can efficiently be implemented in ConvNets by using the translational invariance property of the Gaussian convolution. The displacement of a Gaussian relative to the filter manifests in a shifted convolution result, i.e.,
\begin{align}	
    f \ast G(\boldsymbol{\mu}_{k};\sigma) &= f \ast \mathcal{T}_{\boldsymbol{\mu}_{k}}[G(\sigma)]\\
    & = \mathcal{T}_{\boldsymbol{\mu}_{k}}[f \ast G(\sigma)],
    \label{equ:gauss-trans-invariance}
\end{align}
where $\mathcal{T}_x(g,y) = g(y-x)$ is translation of function $g(\cdot)$ and $G(\sigma)$ is zero-mean Gaussian. Thus the activation map computation can be written as:
\begin{align}
Y_{i}&=f\left(\underset{s}{\sum}\underset{k}{\sum}w_{k}\mathcal{T}_{\boldsymbol{\mu}_{k}}(G(\sigma) \ast X_{s})+b_{s}\right). \label{equ:fast-gauss-cnn}
\end{align} 
This formulation affords an efficient implementation by pre-computing 
convolutions of all inputs by a single Gaussian kernel, i.e., $\tilde{X}_{s}=G(\sigma) \ast X_{s}$, and applying displacements by $\boldsymbol{\mu}_{k}$ to compute the aggregated responses of each output neuron.

Note that due to discretization, Eq.~(\ref{equ:fast-gauss-cnn}) is accurate only for discrete displacements $\boldsymbol{\mu}_{k}$. We address this by re-defining the translation function in Eq.~(\ref{equ:fast-gauss-cnn}) as a bilinear interpolation
\begin{align}
\mathcal{T}_x(g,y) &= \underset{i}{\sum}\underset{j}{\sum} a_{i,j} \cdot g(y - \left \lfloor{x}\right \rfloor + [i,j]),
\end{align}
where $a_{i,j}$ are bilinear interpolation weights. This now allows us to perform sub-pixel displacements and can be efficiently implemented in CUDA kernels.

\subsection{Learning DAU filter}

The DAU contains two learnable parameters: the input amplification $w_{k}$ and the spatial displacement ${\mu}_{k}$. In principle, the shared aggregation perimeter $\sigma$ could be learned as well, but we found that fixing this value was sufficient in our experiments. Thus the hyperparameters in DAU filters are the aggregation perimeter and the number of DAUs per filter.
 
Since DAUs are analytic functions, the filter parameters are fully differentiable and conform with the standard ConvNet gradient-descent learning techniques with backpropagation. The required partial derivatives are 
\begin{align}
\frac{\partial l}{\partial w_{k}}&=\underset{n,m}{\sum}\frac{\partial l}{\partial z}\cdot\underset{\boldsymbol{x}}{\sum}\mathcal{T}_{\boldsymbol{\mu}_{k}} (X_{s} \ast G(\sigma)),\\
\frac{\partial l}{\partial\mu_{k}}&=\underset{n,m}{\sum}\frac{\partial l}{\partial z}\cdot \underset{\boldsymbol{x}}{\sum} w_{k} \cdot \mathcal{T}_{\boldsymbol{\mu}_{k}} (X_{s}\ast\frac{\partial G(\sigma)}{\partial\mu}),
\end{align}
where $\frac{\partial l}{\partial z}$ is back-propagated error. 

Similarly to inference in Sec.~\ref{sec:inference}, the gradient can efficiently be computed using convolution with zero-mean Gaussian (or derivatives) and sampling the response at displacement specified by the mean values in the DAUs. This significantly reduces the computational cost compared to the explicit mixture model filters~\cite{Tabernik2016a}. 
 
The backpropagated error for the lower layer is computed similarly to the classic ConvNets, which convolve the backpropagated error on the layer output with rotated filters. Since the DAUs are rotation symmetric themselves, only the displacements have to be rotated about the origin and Eq.~(\ref{equ:fast-gauss-cnn}) can be applied for computing the back-propagated error as well, yielding efficient and fast computation.
 
\section{Performance evaluation\label{sec:eval}}

This section reports results of the experimental evaluation of DAUs. We first analyze the influence of the hyperparameters on DAU filters and then evaluate our approach by replacing the standard filters in ConvNets with DAUs filters for a classification task (Sec.~\ref{sec:class-perf}) and a segmentation task (Sec.\ref{sec:sem-segment}). Sec.~\ref{sec:dau-analysis} reports analysis of the learned filter receptive fields and how the number of DAUs per filter impacts the ConvNet performance. 

\subsection{Hyperparameter analysis in DAU-ConvNets}
\label{sec:hyperparam}

We analyze the influence of two hyperparameters on our network: (a) variance $\sigma^2$ used in aggregation and (b) the number of DAUs per filter. We analyzed both on classification problem using CIFAR10~\cite{Krizhevsky2009} dataset. 

For the purpose of this evaluation we used a shallow network with only three convolutional layers with DAU filters and three max-pooling layers. To classify the whole image we appended fully-connected layer. Batch normalization~\cite{He2014} was applied to convolutional layers and weights were initialized using~\cite{Glorot2010}.  We trained the network with softmax loss function for 100 epochs using a batch size of 256 images. Learning rate was set to 0.01 for the first 75 epochs and reduced to 0.001 for remaining epochs. We used momentum of 0.9 as well.
\vspace{-1em}
\paragraph{Variance:} When evaluating variance we fixed the number of DAUs per filter to four and varied the variance $\sigma^2$ from $0.3^2$ to $0.8^2$. Results are reported in Tab.~\ref{tab:variance-cifar}. They indicate that the variances have negligible effect on classification performance with changes between different variances at only around 1\%. We used variance of $\sigma^2=0.5^2$ for all remaining experiments in this paper.
\vspace{-1em}
\paragraph{Number of DAUs per filter:} When evaluating the number of units we used a variance $\sigma^2=0.5^2$ and varied the number of units on the second and the third layer using 1, 2, 4 or 6 units. We fixed DAUs on the first layer to four units  to capture initial edges and corners. Results are reported in Tab.~\ref{tab:num-units-cifar}. They indicate only a slight increase of performance when additional units are added. Difference between using a single unit or using six units is only 1\%. We used two and four units in remaining experiments as a trade-off between performance and parameter count. Additional extensive evaluation of parameter count was perform in Sec.~\ref{sec:param-analysis}.

\subsection{Classification performance\label{sec:class-perf}}

\begin{table}
\centering
\small
\caption{Variance $\sigma^2$ hyperparameter evaluation on CIFAR10 classification task using a shallow DAU-ConvNet. Variance has minor effect on classification performance.}
\label{tab:variance-cifar}
\vspace*{-0.5em}
\begin{adjustbox}{width=\columnwidth}

\begin{tabular}{cp{15pt}p{15pt}p{15pt}p{15pt}p{15pt}p{15pt}}
\hline
Variance $\sigma^2$ & $0.3^2$ & $0.4^2$ & $0.5^2$ & $0.6^2$ & $0.7^2$ & $0.8^2$ \\
\hline
\hline
DAU-ConvNet  & \multirow{2}{*}{82.9} & \multirow{2}{*}{83.4} & \multirow{2}{*}{\textbf{83.8}} & \multirow{2}{*}{83.6} & \multirow{2}{*}{82.9} & \multirow{2}{*}{82.8} \\
 CIFAR10 & & & & & & \\
 \hline
\end{tabular}
\end{adjustbox}
\end{table}

Performance of DAUs on the classification task was tested on the ILSVRC 2012 dataset~\cite{Russakovsky2015}. A standard testing protocol was used. The network was trained on 1.2 million images and tested on the validation set with 50,000 images. All images were cropped and resized to 227 pixels. To keep the experiments as clean as possible, we did not apply any advanced augmentation techniques apart from mirroring during the training with probability $0.5$.
 
As our baseline ConvNet architecture, we chose the AlexNet model~\cite{Krizhevsky2012}, which is composed of 7 layers: 5 convolutional and 2 fully connected. We retained the local normalization layers, max-pooling and dropout on fully-connected layers of the original AlexNet~\cite{Krizhevsky2012}, but we did not split channels into two streams as was done in the original work~\cite{Krizhevsky2012}. We also used weight initialization technique by Glorot and Bengio~\cite{Glorot2010}.

The baseline ConvNet was modified into a DAU-ConvNet as follows. The filters in the convolutional layers from layer 2 to 5 were replaced by our DAU filters from Sec.~\ref{sec:method}. Four DAUs per filter were used in the second layer and two DAUs per per filter in the remaining three layers. This follows approximate coverage of filter sizes from classic ConvNet with $5\times5$ filter sizes for the second layer and $3\times3$ filter sizes for the remaining layers. First layer and fully connected layers remained unchanged using classic convolutional layer. This is partially due to technical limitation of our current implementation. Our recent work on alleviating this issues indicates that even fully connected layers with 36 units ($6\times 6$ filter sizes) can be replaced with only 6 DAUs (with comperable perfomance).

\begin{table}
\centering
\small
\caption{Number of units per filter hyperparameter evaluation on CIFAR10 classification task using a shallow DAU-ConvNet. Larger number of units increase classification performance only slightly.
}
\label{tab:num-units-cifar}
\vspace*{-0.5em}
\begin{adjustbox}{width=0.9\columnwidth}
\begin{tabular}{ccccc}
\hline
Number of units  per filter & 1 & 2 & 4 & 6\\
\hline
\hline
DAU-ConvNet  & \multirow{2}{*}{82.9} & \multirow{2}{*}{83.3} & \multirow{2}{*}{83.8} & \multirow{2}{*}{\textbf{84.1}}  \\
 CIFAR10 & & & &  \\
 \hline
\end{tabular}
\end{adjustbox}
\vspace*{-1em}
\end{table}
\subsubsection{Optimization}
We trained both, ConvNet as well as DAU-ConvNet, with stochastic gradient descent using batch size of 128. Both models were trained for 800,000 iterations, or 80 epochs, with initial learning rate of 0.01, which is reduced by a factor of 10 every 200,000th iteration. We used momentum with a factor of 0.9 and a weight decay factor of 0.0005. In our layers with DAUs a decay factor could be applied to weights and offsets as well, although applying to offsets has a different effect than decay on regular weights as it would prevent them from moving further from the center. We used decay only on weights but not on the offsets.

\subsubsection{Classification results}

\begin{figure}
\centering
\includegraphics[width=0.9\columnwidth]{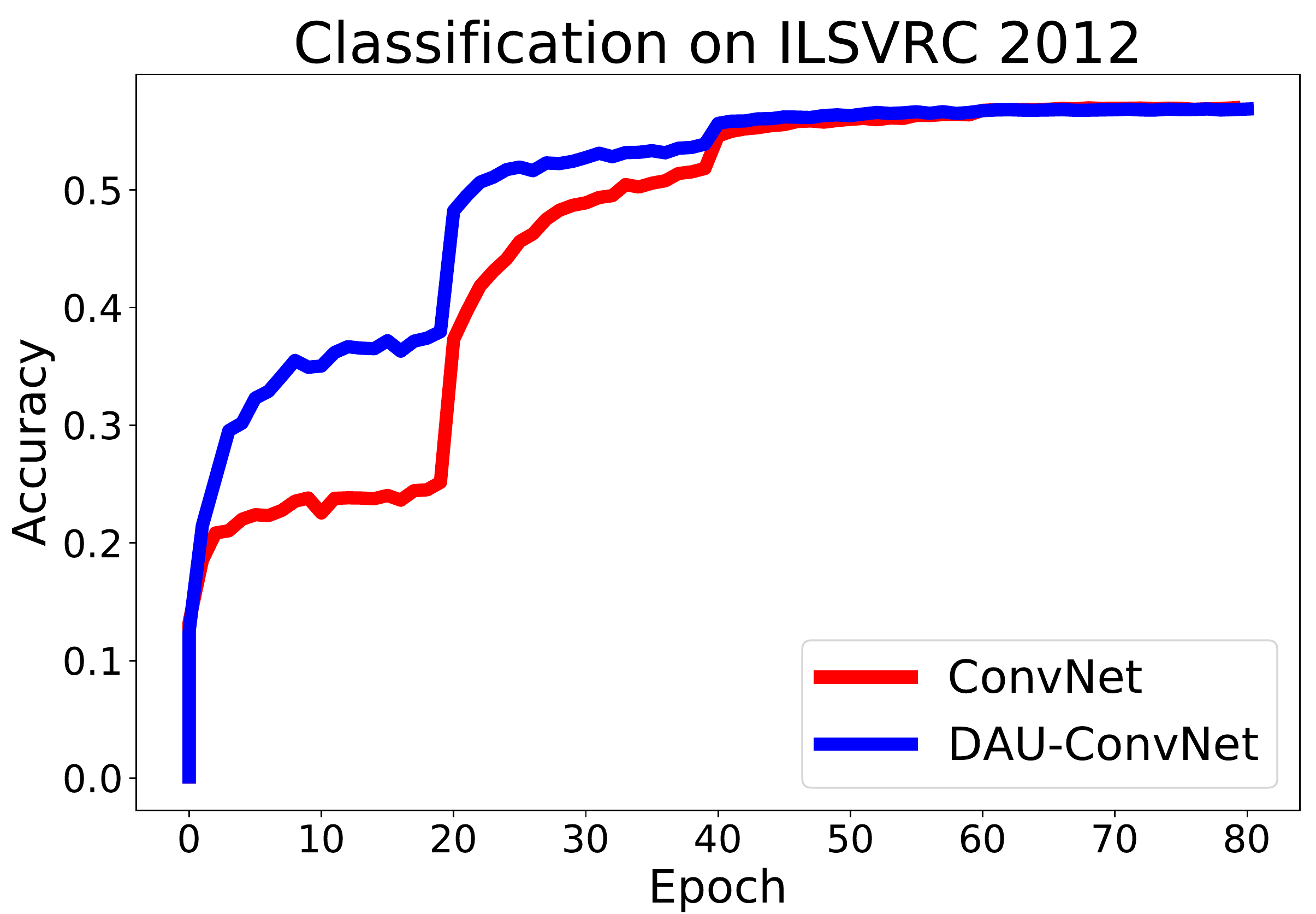}
\vspace*{-0.5em}
\caption{Classification top-1 accuracy on ILSVRC 2012 validation set using AlexNet architecture. Our DAU-ConvNet converges faster with larger learning rates than standard ConvNet.
\label{fig:cls-results}}
\end{figure}

The results are reported in Tab.~\ref{tab:cls-results} with performance monitoring during the training reported in Fig.~\ref{fig:cls-results}. After 600,000 iterations, the DAU-ConvNet and the baseline ConvNet converge to a comparable performance. Namely, after 80 epochs, both models achieved accuracy of slightly below 57\% (see Tab.~\ref{tab:cls-results}), however, DAU-ConvNet was converging much faster, resulting in higher performance jumps before the learning rate reduction steps. Tab.~\ref{tab:cls-results} shows the number of free parameters in the convolutional layers. Note that DAU-ConvNet requires 30\% less parameters than the baseline classic ConvNet and our analysis in Sec.~\ref{sec:param-analysis} shows this can be improved even further.

Even though the final classification performance converges to the same result, our model exhibits good performance even on higher learning rates. This indicates that the DAUs modify landscape of the loss function so that it can be traversed faster with higher learning rates in DAU-ConvNets. This improvement may also be contributed to reduction of the number of parameters in the DAUs and supports the hypothesis that DAUs do not lose expressive power on the account of their simple functional form.

\subsection{Semantic segmentation\label{sec:sem-segment}}

We analyze the performance of DAUs on a dense prediction problem where large receptive fields and fine resolution are particularly important.
In this experiment, we start from the baseline ConvNet and DAU-ConvNet trained in Sec.~\ref{sec:class-perf} and fine-tune them for a segmentation task. A standard technique is used to modify the classification networks into segmentation nets. Specifically, the last fully-connected classification layer is replaced by the expansion and classification layer from Long et al.~\cite{Long2015} that entails a $1\times1$ classification layer and up-sampling using a deconvolution layer to obtain pixel-wise loss. To keep the experiments clean we have not added advanced network adaptations that have emerged over recent years, like feature combination across layers, etc., although our approach is general enough to allow such upgrades. The object boundaries are maintained sharp by further increasing higher layer resolution.

\subsubsection{Increasing resolution at higher layers}
Our classifier from Sec.~\ref{sec:class-perf} follows the AlexNet architecture and reduces the resolution by 32-fold. For the purpose of segmentation we increase the resolution at higher layers and remove the last two max-pooling layers thus reducing resolution only by 8-fold for segmentation. With this modification, the network retains finer details. 

Increasing the resolution on a pre-trained model causes a misalignment of already learned filter weights and their positions w.r.t. the expected resolution. We compensate for that by modifying the parameters of the affected layers. In particular, for the layers with DAU filters we increase displacement of a unit with the appropriate factor, while in classic ConvNet layer we use dilated convolution with the same factor. The layers after the first-removed-max-pooling use a factor of two (layers 3--5) and the layers after the second-removed-max-pooling use a factor of four (layer 6). 

\subsubsection{Dataset}
We evaluate our segmentation DAU-ConvNet and the baseline ConvNet with dilation on PASCAL VOC 2011 segmentation dataset. For the training we use 1,112 training images from PASCAL VOC 2011 segmentation combined with 7,386 images collected by Hariharan et al.~\cite{Hariharan2011}. We report results on PASCAL VOC 2011 validation set excluding the images from~\cite{Hariharan2011} that were also used for training. 

\subsubsection{Optimization}
We trained the models with mini-batch stochastic gradient descent and a batch size of 20 images for 65,000 iterations, or 150 epoch. We used a fixed learning rate of 0.0002, weight decay of 0.0005 and momentum of 0.9. The added classification layer was initialized with zeros, similar to~\cite{Long2015} and we used a normalized per-pixel softmax loss function applied only to pixels with a valid annotation. 

\subsubsection{Segmentation results}

\begin{table}
\centering
\caption{Results on ILSVRC 2012 validation set using AlexNet architecture and corresponding number of parameters on convolutional layers. We report top-1 accuracy. \label{tab:cls-results}}
\vspace*{-0.5em}
\begin{tabular}{lccc}
\hline
 &  Top-1 & Number of parameters  \\ 
 &  accuracy (\%) & on conv. layers\\
\hline
\hline
DAU-ConvNet & 56.89  &  2.3 mio  \\
ConvNet~\cite{Krizhevsky2012} & 56.99 & 3.7 mio \\
\hline
\end{tabular}
\end{table}
\begin{figure*}[t]
\includegraphics[width=0.33\textwidth]{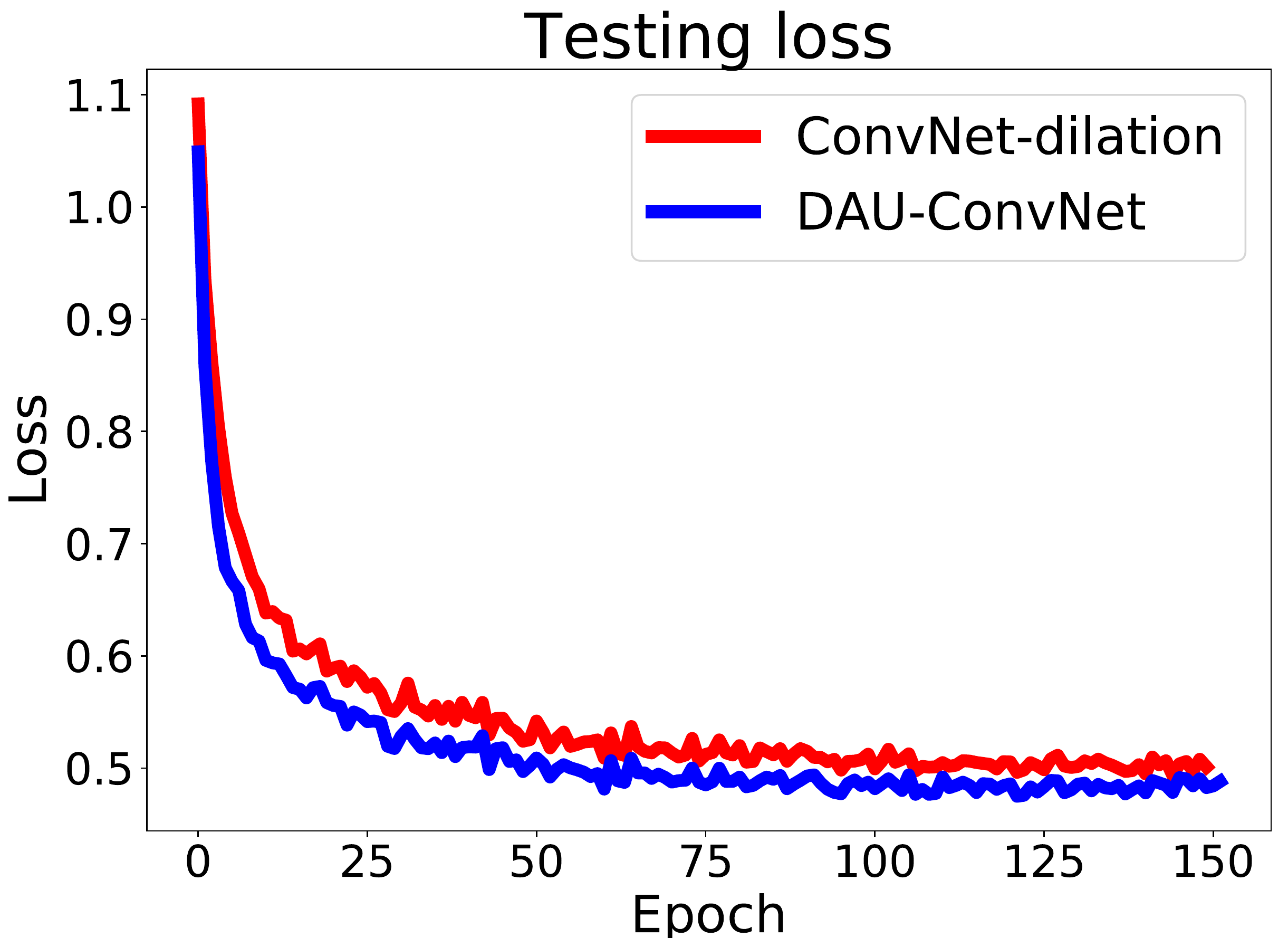}\includegraphics[width=0.33\textwidth]{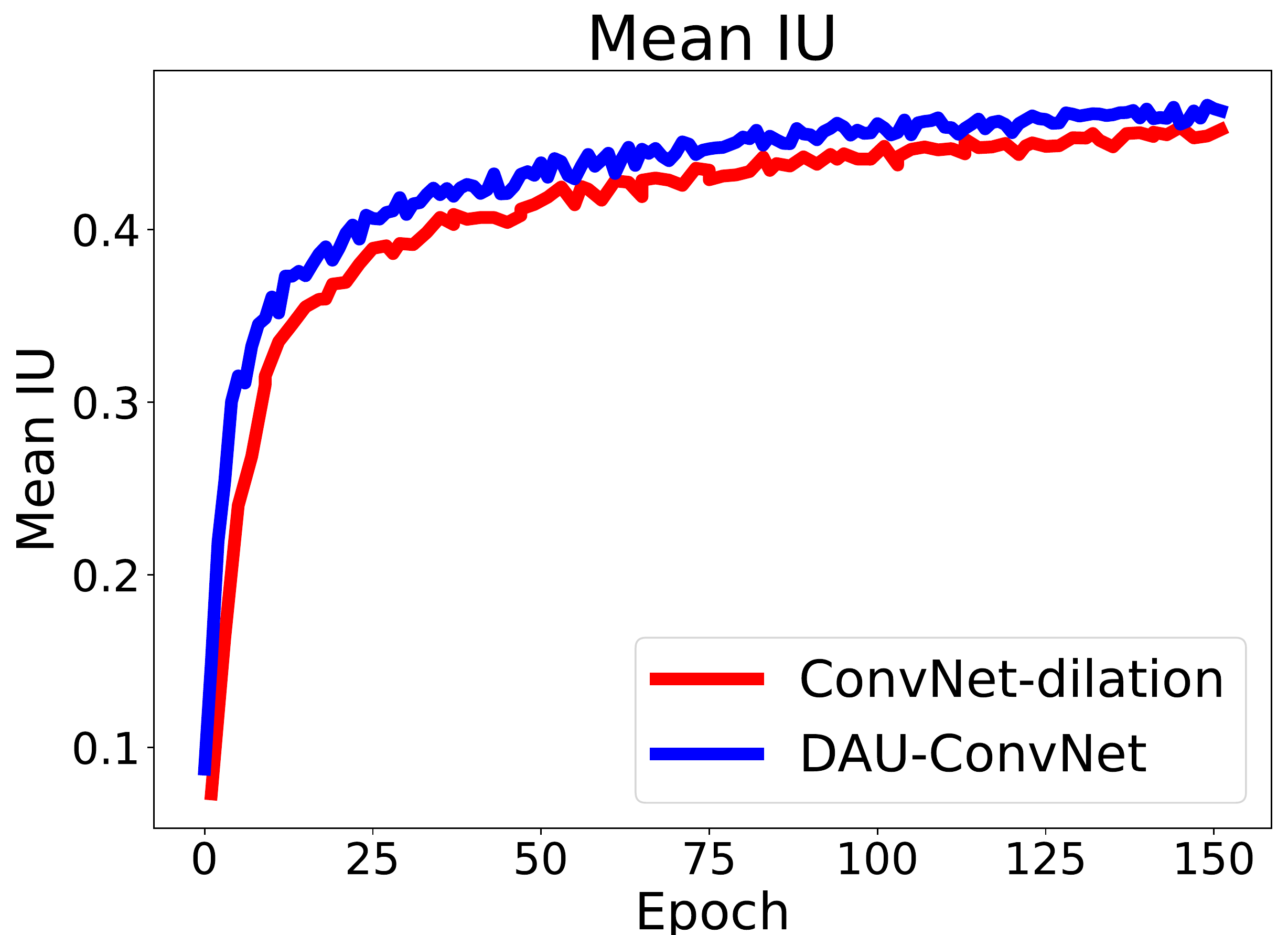}\includegraphics[width=0.33\textwidth]{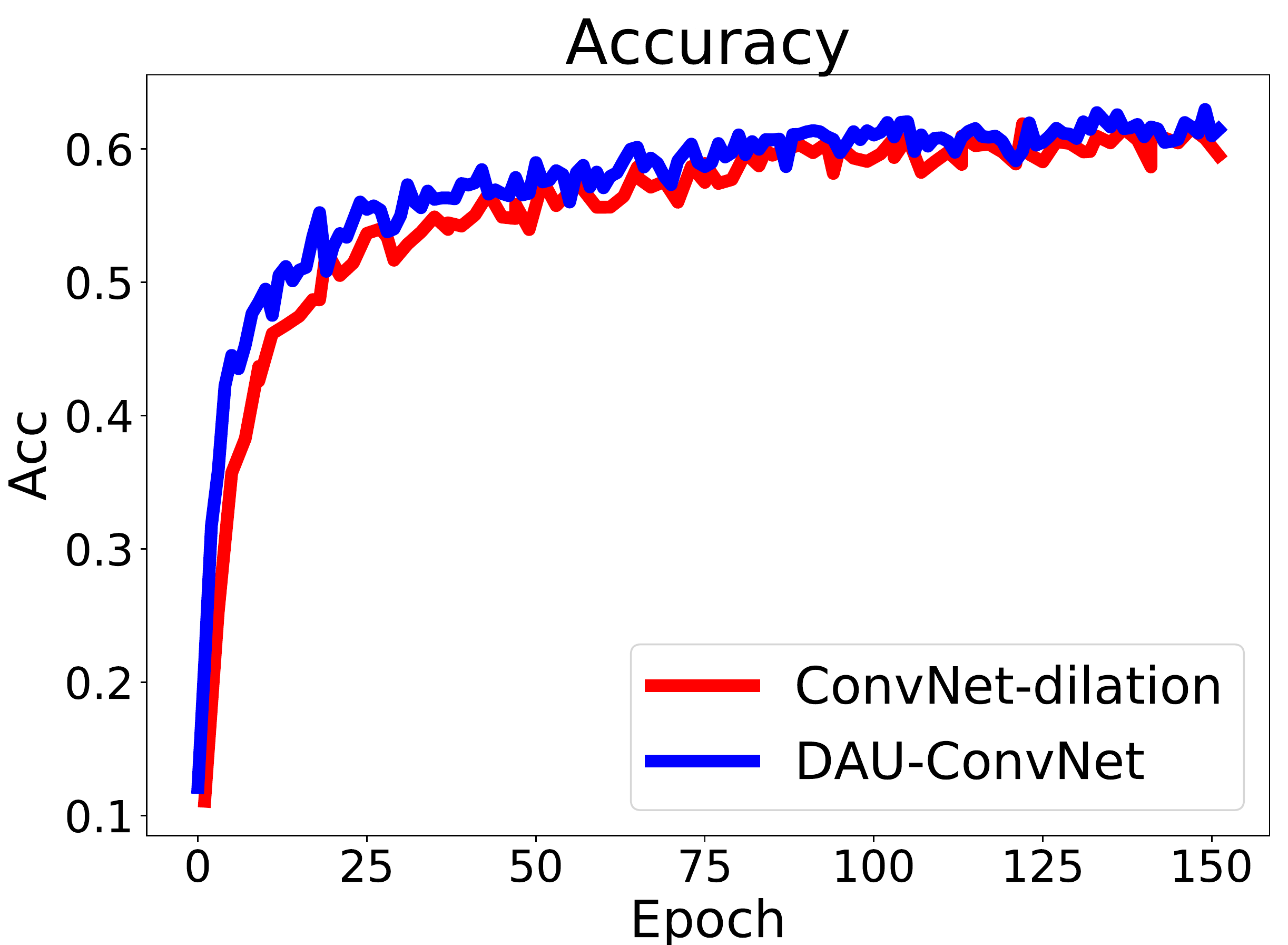}
\vspace*{-0.5em}
\caption{Performance monitoring during fine-tuning on segmentation task. Results are reported on PASCL VOC 2011 segmentation validation set. We report testing loss value and averaged mean-iu and accuracy.\label{fig:seg-results}}
\end{figure*}

\begin{table*}
\small{
\caption{Results on segmentation task using a PASCAL VOC 2011 validation set. We report per-class mean-IU and averaged mean-IU over all classes.  \label{tab:voc2011-results}}
\vspace*{-1em}
\begin{adjustbox}{width=1\textwidth}
\begin{tabular}{ 
c|p{0.3cm}p{0.3cm}p{0.3cm}p{0.3cm}p{0.3cm}p{0.3cm}p{0.3cm}p{0.3cm}p{0.3cm}p{0.3cm}p{0.3cm}p{0.3cm}p{0.3cm}p{0.3cm}p{0.3cm}p{0.3cm}p{0.3cm}p{0.3cm}p{0.3cm}p{0.3cm}c|c }
\hline
& \rot{background} & \rot{aeroplane} & \rot{bicycle} & \rot{bird} & \rot{boat} & \rot{bottle} & \rot{bus} & \rot{car} & \rot{cat} & \rot{chair} & \rot{cow} & \rot{diningtable} & \rot{dog} & \rot{horse} & \rot{motorbike} & \rot{person} & \rot{potted plant} & \rot{sheep} & \rot{sofa} & \rot{train}&\rot{tv/monitor} & mean IU\\
\hline
\hline

DAU-ConvNet & \textbf{86.1} & \textbf{58.5} & \textbf{29.7}& \textbf{55.0} & \textbf{41.7} & \textbf{47.2} & \textbf{61.3} & \textbf{56.3}& \textbf{57.9} & \textbf{14.1} & \textbf{47.1} & \textbf{27.3} & 47.8 & \textbf{36.7} & \textbf{54.7} & \textbf{63.9} & \textbf{28.9} & 53.0 & \textbf{19.3} & 59.8 & \textbf{45.3} & \textbf{47.22}\\
ConvNet-dilation & 85.8 & 54.6 & 27.2 & 51.8 & 39.0 & 45.2 & 56.3 & 54.2 & 57.4 & 12.4 & 43.8 & 26.1 & \textbf{50.6} & 35.6 & 54.1 & 61.1 & 26.9 & \textbf{53.6} & 18.9 & \textbf{60.2} & 42.5 & 45.57\\ 
\hline
\end{tabular}

\end{adjustbox}
}
\end{table*}

The performance of DAU-ConvNet compared to the baseline ConvNet with dilation is shown in Fig.~\ref{fig:seg-results}. The DAU-ConvNet shows faster convergence in testing loss. In addition, DAU-ConvNet shows consistently better segmentation performance than the baseline ConvNet-dilation across all measures. The mean IU and per-pixel accuracy are improved by approximately 2\%. Looking at the per-class mean IU in Tab.~\ref{tab:voc2011-results}, we observe improved performance across all categories, with the exception of "dog", "sheep" and "train". 
 

\section{Analysis of displaced aggregation units\label{sec:dau-analysis}}

\begin{figure*}
\begin{subfigure}[]{0.33\textwidth}
\includegraphics[width=\textwidth]{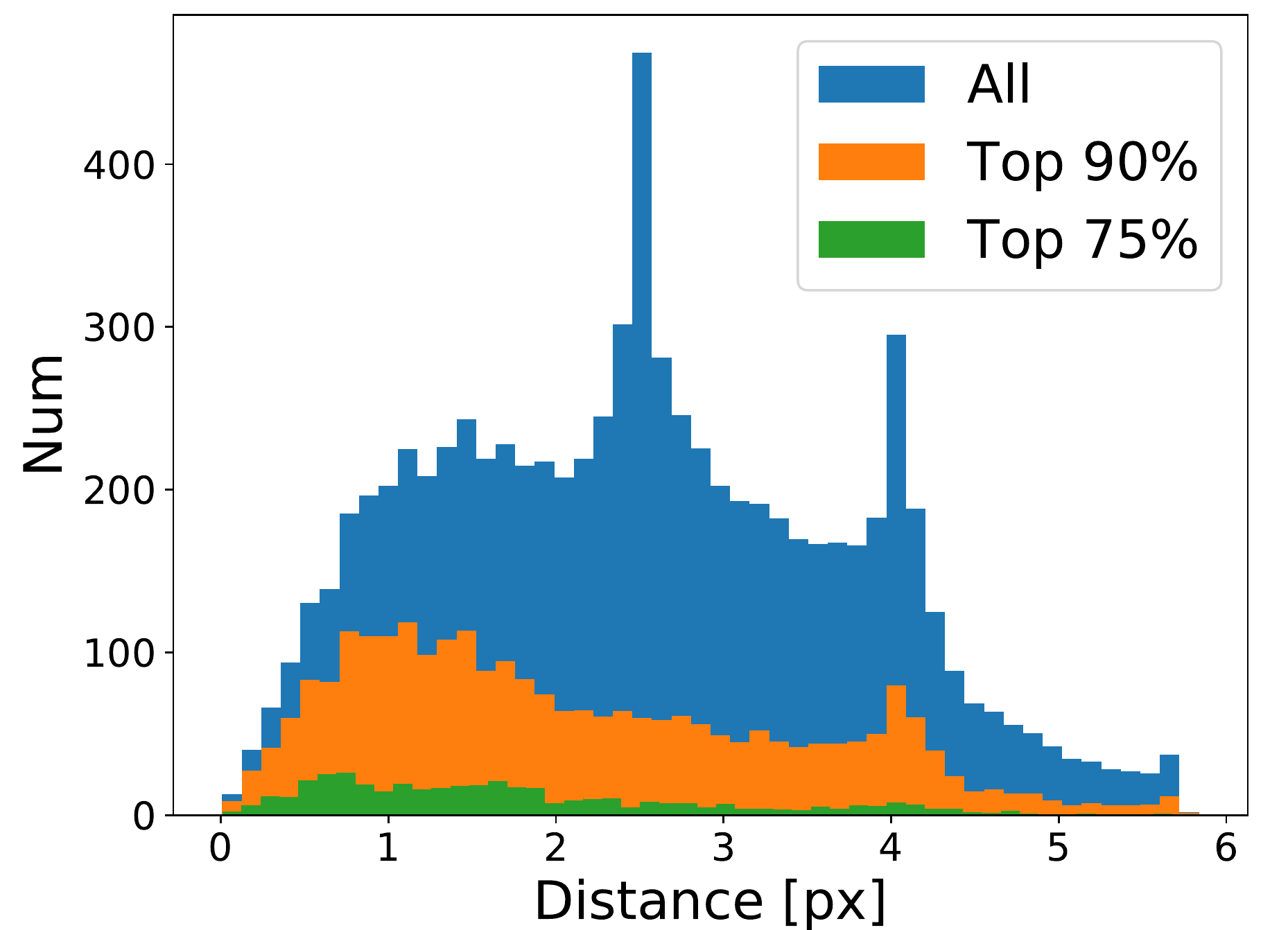}
\caption{Layer 3\label{fig:spatial-dist-1d-ly3}}

\end{subfigure}\begin{subfigure}[]{0.33\textwidth}
\includegraphics[width=\textwidth]{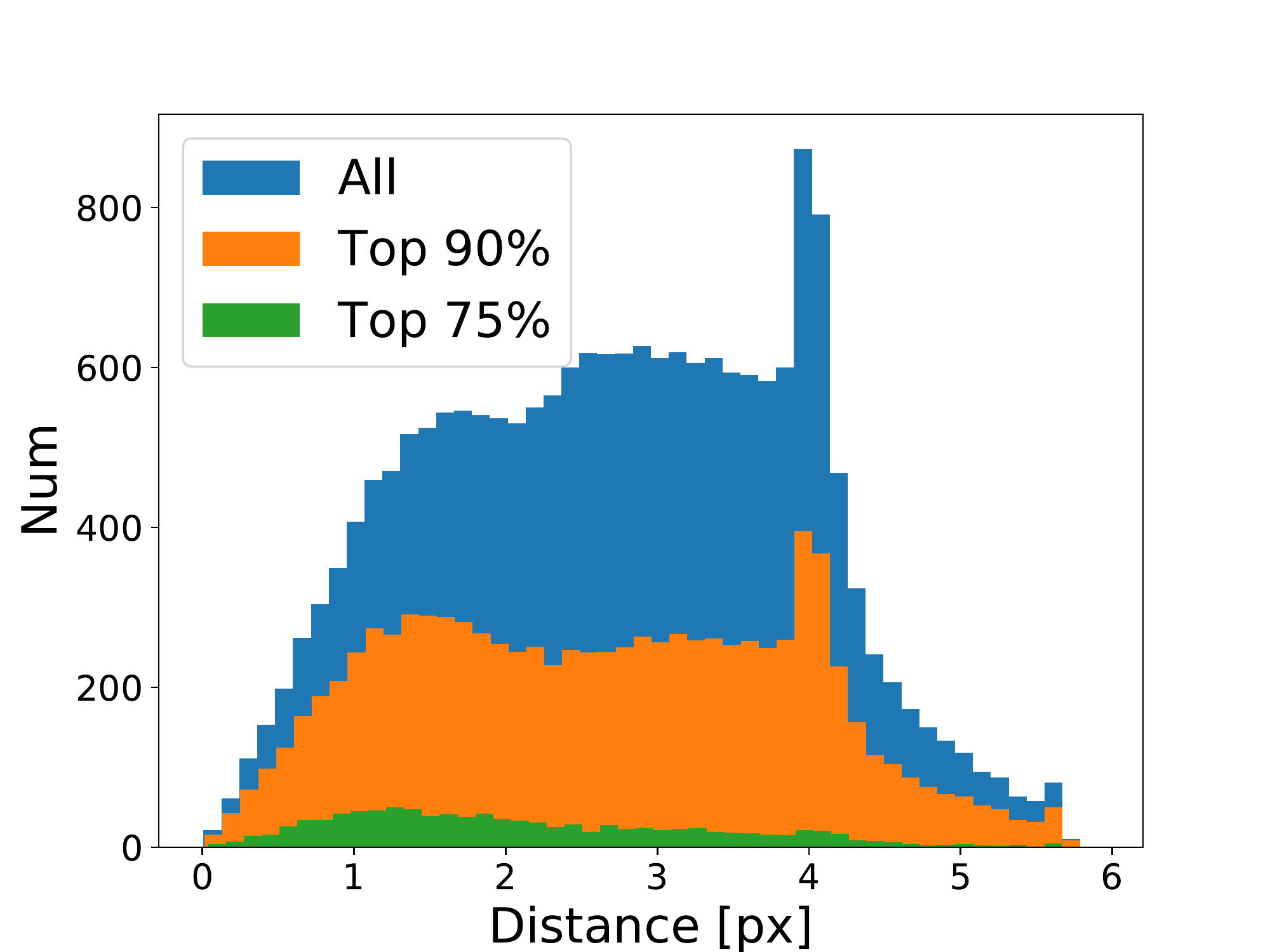}
\caption{Layer 4\label{fig:spatial-dist-1d-ly4}}
\end{subfigure}\begin{subfigure}[]{0.33\textwidth}
\includegraphics[width=\textwidth]{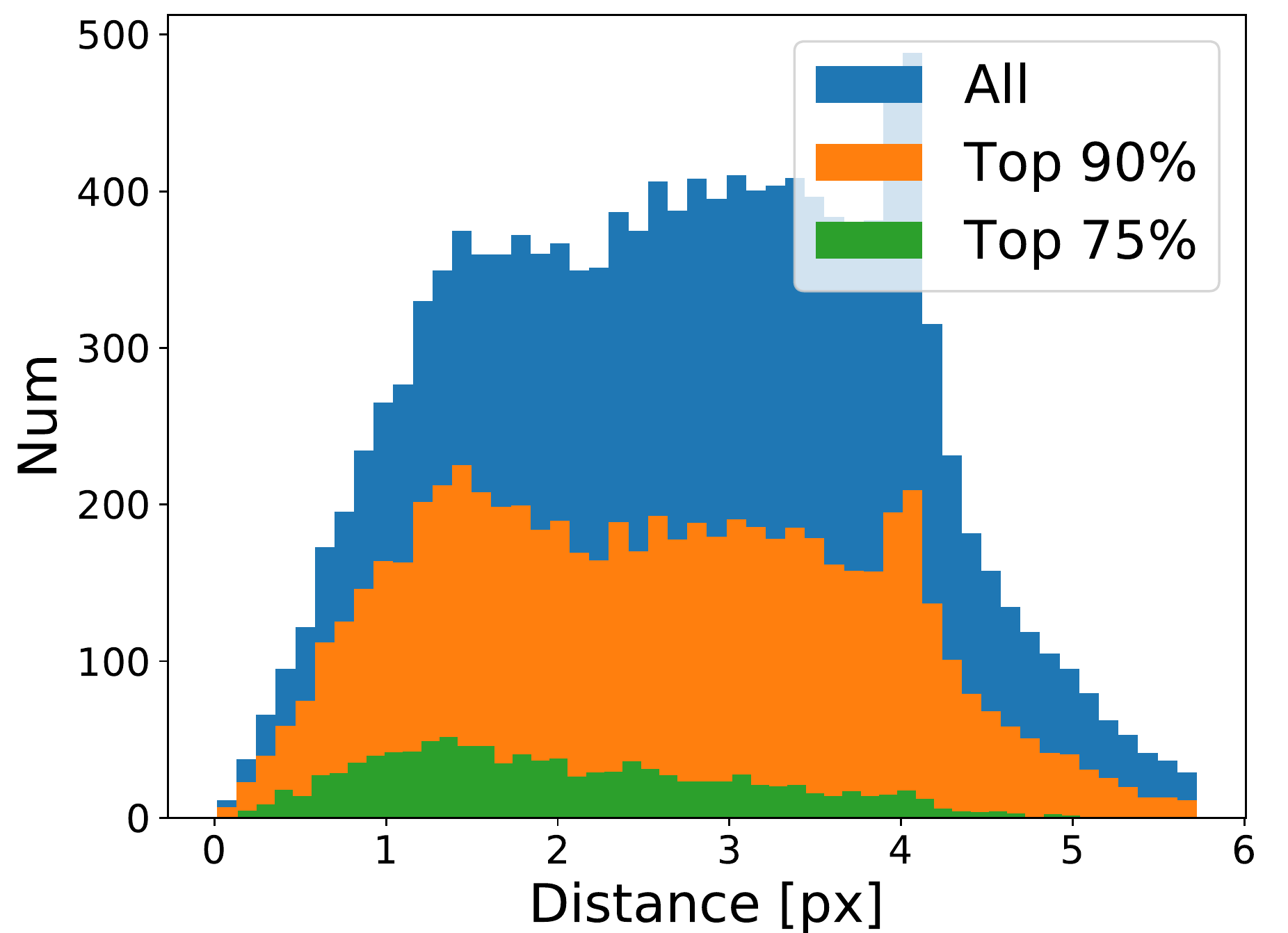}
\caption{Layer 5\label{fig:spatial-dist-1d-ly5}}
\end{subfigure}
\vspace*{-0.5em}
\caption{Distance-to-center distributions collected from displacement of DAUs. Distributions reported per-layer (columns) and after elimination of units with smallest amplification factor using different relative thresholds (colors).\label{fig:spatial-dist-1d}}
\end{figure*}

In this section we conducted two experiments to gain further insights into DAUs. The first experiment analyzed the spatial distribution of the DAUs in the learned filters (Sec.~\ref{sec:adaptation-analysis}). The second experiment explored the relation between the number of DAUs per filter and the network performance (Sec.~\ref{sec:param-analysis}).


\subsection{Spatial adaptation of filter units\label{sec:adaptation-analysis}}

We investigate spatial distribution of DAUs in our network by observing distributions of the learned displacements in the segmentation DAU-ConvNet in Sec.~\ref{sec:sem-segment}. The aim of the experiment was to expose two aspects: (i) the distribution of the learned displacements, which indicates displacement locations favored for a given task, and (ii) overall spatial distribution, which indicates the preferred receptive field size. 

Such an experiment is very difficult to perform with classical ConvNets and requires a combinatorial sweep over alternative architectures with various manually-defined filter designs. For example, dilated convolutions can alter unit positions, but this must be done with a specific pre-defined dilation factor. In contrast, with displaced aggregated units in our filters we can analyze their displacements that adjust during the learning on the segmentation problem with a sub-pixel accuracy and not being confined to the same pattern across all filters. Such an analysis is not possible with the existing ConvNet architectures.


\begin{figure}
\includegraphics[width=\columnwidth]{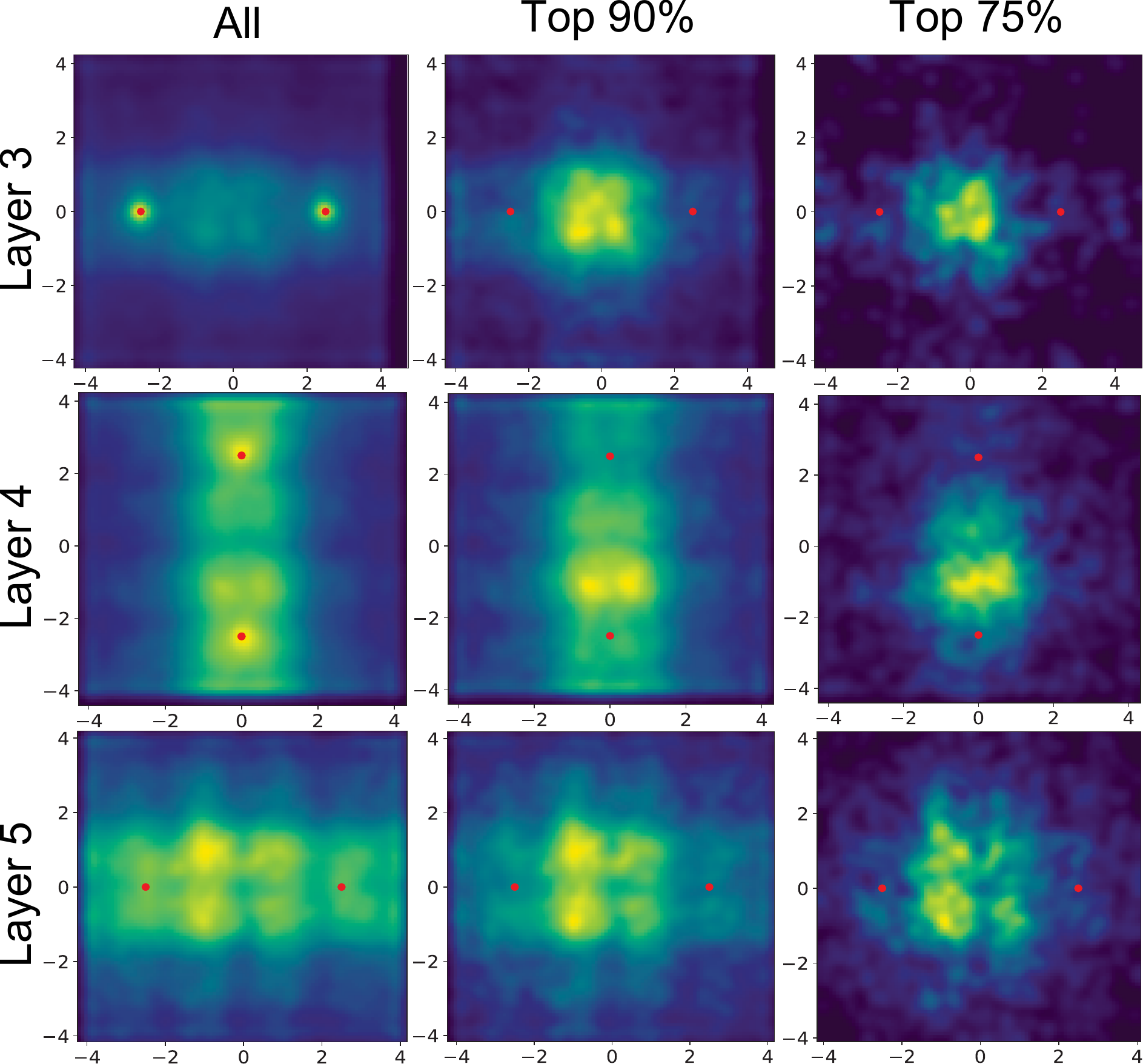}
\caption{2D distributions of displacements collected from DAUs. Red dots indicate initialization points. Distributions reported for layer 3, 4 and 5 in top, middle and bottom row, respectively, and each after retaining different number of important units (in columns).\label{fig:spatial-dist-2d}}
\end{figure}

We investigate two types of distributions: (i) a 1D distance-to-center distribution and (ii) a distribution of displacements in 2D space. We obtain 1D spatial distribution by collecting displacement values of units from all features at a specific layer and compute their distances to the center of the filter. All distances are collected in a histogram with each unit contributing with its corresponding input amplification factor. We obtain the second 2D spatial distribution by plotting all displacements from a specific layer into the same graph. 

\subsubsection{Results and discussion}

We compute several per-layer distributions from the DAU-ConvNet model trained for semantic segmentation in Sec.~\ref{sec:sem-segment}. All 1D distributions are shown in Fig.~\ref{fig:spatial-dist-1d} and all 2D distributions are shown in Fig.~\ref{fig:spatial-dist-2d}. The first set of distributions is computed from all absolute amplification values $|w_k|$. The second distribution is obtained by retaining $90\%$ of the largest absolute amplification values and the third distribution by retaining $75\%$ of the largest absolute amplification values.


We observe in Fig.~\ref{fig:spatial-dist-1d} two significant spikes, one at 2.5 pixels and another at 4 pixels away from the center. The spike at 2.5 pixels, that occurs only at the third layer, is artificial due to the fixed initialization points. It indicates that many units did not move from their initialization point during learning. This can be observed in Fig.~\ref{fig:spatial-dist-2d} with high density at initialization center points (red dots). Further inspection shows that those units do not contribute to the filter significantly. In fact, they disappear in the plots when units with lowest amplification value are removed.

The 4th and the 5th layers have similar initialization points but no apparent spikes in their distance-to-center distributions, as shown in (Fig.~\ref{fig:spatial-dist-1d-ly4} and Fig.~\ref{fig:spatial-dist-1d-ly5}). This indicates a low learning rate for the 3th layer where displacements may not have been able to move quickly enough. As results in Sec.~\ref{sec:param-analysis} suggest, some of these may be removed without performance reduction. 


The second spike at 4 pixels away from the center (Fig.~\ref{fig:spatial-dist-1d}) is more significant since it does not disappear when removing units with small amplification factors. This spike occurs due to an artificial limit on boundary of the receptive field which in our case is set at four pixels in both spatial dimensions\footnote{Our current Caffe/CUDA implementation allows distances only up to 4 or 8 pixels. This can be overcome with a improved implementation.}. Still, a significant number of those units have large amplification factors. This points to the need of further increasing the allowed sizes of the receptive fields. 

A consistent shape of distance-to-center distributions throughout the layers (Fig.~\ref{fig:spatial-dist-1d}) points to a desired spatial distribution of units for segmentation. It indicates that units must densely cover locations at a distance of 1-2 pixels away from the center Some units with high amplification factor are located far away from the center which indicates a need to covering larger receptive fields albeit with lower density. The same conclusion is drawn from 2D spatial distributions in Fig.~\ref{fig:spatial-dist-2d}.

\subsection{Parameter-space analysis\label{sec:param-analysis}}
 
\begin{table*}
\centering
\small{
\caption{Analysis of the number of parameters and units per filter with three variants of DAU-ConvNet: Large, Medium and Small. Rows also show the elimination of units based on their amplification value. In columns we report classification top-1 accuracy on ILSVRC2012 validation set, the number of DAU on all filters and percentage of removed units. \label{tab:paramater-study}}

\begin{adjustbox}{width=1\textwidth}
\begin{tabular}{r|ccc|ccc|ccc}
\hline
Relative & \multicolumn{3}{c|}{Large DAU-ConvNet} & \multicolumn{3}{c|}{Medium DAU-ConvNet} & \multicolumn{3}{c}{Small DAU-ConvNet}\\
threshold & Acc. (\%) & \# units & \% removed  & Acc. (\%) & \# units & \% removed & Acc. (\%) & \# units & \% removed \\ 
\hline
\hline
0 & \textbf{57.3} & 1,523,712 & 0 & 56.9 &  786,432 & 0 & 56.4 & 393,216 & 0  \\
0.01 & \textbf{57.3} & 1,389,131 & 8  & 56.8 & 739,884 & 6 & 56.4 & 378,692 & 4 \\
0.02 & 57.1 & 1,325,057 & 13 & 56.7 & 707,745 & 10 & 56.4 & 366,144 & 7 \\
0.05 & 40.1 & 1,157,129 & 24 & 54.8 & 623,923 & 20 & 55.4 & 331,137 & 16 \\
0.10 & 28.3 & 925,509 &  39 & 47.4 & 507,651 & 35 & 49.6 & 279,162 & 29\\
0.25  & 0.2 & 453,987 & 70 & 1.9 & 261,093 & 66 & 0.9 & 154,624 & 61 \\
\hline
\end{tabular}
\end{adjustbox}
}
\end{table*}
We analyzed the impact of the number of DAUs per filter on the network performance to gain additional insights. Several research papers investigate the influence of parameter space in classic ConvNets with respect to the number of layers, number of features or filter sizes~\cite{Eigen}, but could not report analysis with respect to the filter units. The classic ConvNets are limited by a minimal filter size of $3\times3$ that already has a minimal spatial coverage. Reducing the parameter count by reducing the filter size would not be feasible. Our redefinition of filter units on the other hand allows us to investigate filters with even smaller number of parameters without affecting spatial coverage and the receptive field sizes.
 

The number of units per filter was set through a hyperparameter. Thus the number of parameters was kept equal across all filters during training. Then the units with small amplification weights were removed

\begin{table}
\centering
\small{
\caption{Per-filter unit and parameter count with three variants of DAU-ConvNet: Large, Medium and Small. Note, a unit in DAU has three parameters and ConvNet has one.\label{tab:paramter-count}}

\begin{tabular}{lcccc}
\hline
	& \multicolumn{4}{c}{Per-filter unit count} \\

	& Large & Medium & Small & ConvNet\\
\hline
\hline
Layer 2 & 6 & 4 & 2 & $5\times5$\\
Layers 3-5 & 4 & 2 & 1 & $3\times3$\\
\hline
	& \multicolumn{4}{c}{Per-filter parameter count}  \\
\hline
\hline
Layer 2 & 18 & 12 & 6 & 25 \\
Layers 3-5 & 12 & 6 & 3 & 9 \\
\hline
\end{tabular}

}
\end{table}

We perform the experiments on a classification problem with ILSVRC 2012 and AlexNet architecture as presented in Sec.~\ref{sec:class-perf}. We used the same optimization settings for all variants.

Three variations of our network are compared (see Tab.~\ref{tab:paramter-count}): Large, Medium and Small. The Medium DAU-ConvNet is the network from Sec.~\ref{sec:class-perf}. The Small DAU-ConvNet uses as few as two or a single DAU per filter, while the Large DAU-ConvNet uses six to four DAUs. This affects the number of learned parameters as follows. The Small DAU-ConvNet contains 400,000 DAUs, the Medium DAU-ConvNet contains 800,000 DAUs, and the Large DAU-ConvNet contains 1.5 mio DAUs. These values translate to 4.5 mio, 2.3 mio, and 1.2 mio parameters on convolutional layers for Small, Medium and Large DAU-ConvNet, respectively. For the reference, the baseline ConvNet from Sec.~\ref{sec:class-perf} contained 3.7 mio units on conv. layers.


    
\subsubsection{Results and discussion}    

The results are reported in Tab.~\ref{tab:paramater-study}. We observe that all three networks achieve classification accuracy of approximately 56-57\% on ILSVRC 2012. These results indicate that DAU-ConvNets may require only one to two units per filter resulting in 3 to 6 parameters per filter on convolutional layers. This is significantly lower than classic networks that already contain 9 parameters for the smallest filter (i.e., $3 \times 3$) and 25 for a moderately large (i.e., $5 \times 5$). The low parametrization is possible in DAU-ConvNets since the network learns on its own the receptive field perimeter without the need to increase the parameter space to cover large displacements.

Furthermore, looking at the performance when eliminating units with small amplification factor reveals further improvements. In all three networks we were able to eliminate 7-13\% of units without affecting their classification performance at all. 



\section{Discussion and conclusion~\label{sec:conclusion}}

We proposed a displaced aggregation filter units (DAUs) to replace a fixed, grid-based unit in existing convolutional networks. The DAUs modify only the convolutional layer in standard ConvNets, but afford several advancements. The receptive field is now learned. The learning is efficient since DAUs decouple the number of parameters from the receptive field size and efficiently allocate the free parameters. We demonstrated this on the classification and segmentation tasks, and showed faster convergence on the classification task and improved performance on the segmentation task.

The DAUs remove the filter size hyperparameter, but introduce a hyperparameter on the DAU's aggregation perimeter size and the number of DAUs per filter. We experimentally showed that both have minor affect on the classification performance. We can set aggregation perimeter size to a fixed value, while a larger number of units per filter marginally increases performance. With less than 1\% drop in performance we can use only one unit per filter. This is a highly interesting result as it suggests that efficient ConvNets can be implemented by replacing general convolution layers by Gaussian filters and a single sub-pixel sampling per filter.

The analysis of learned DAU displacements showed that units are concentrated at the filter center, while some are positioned further away. This shows the capacity to learn small as well as large receptive fields within a unified framework. Our distributions directly point to locations that need to be densely sampled in filters. This can be used to adjust the dilation factors from atrous convolutions in classical ConvNets~\cite{Chen2017} more efficiently. 

\let\thefootnote\relax\footnotetext{
\textbf{Acknowledgments.} This work was supported in part by the following research projects and programs: project GOSTOP C3330-16-529000 and ViAMaRo L2-6765, program P2-0214 financed by Slovenian Research Agency ARRS, and MURI project financed by MoD/Dstl and EPSRC through EP/N019415/1 grant.
}

Lastly, our comprehensive study of per-filter parameter allocation showed an inefficient allocation of parameters in existing ConvNets. DAU-ConvNets achieved comparable performance to classic CovnNets at 3-times less parameters per filter. Analysis shows there is also room for further improvements as elimination of units with lowest amplification factors (even without post-hoc fine-tunning) can save 10\% of parameters without sacrificing the performance. Furthermore, our recent preliminary work on applying DAUs to fully connected layers indicates possible savings in parameters for fully connected layers as well.

{\small
\bibliographystyle{ieee}
\bibliography{library}
}

\end{document}